\documentclass[letterpaper, 10 pt, conference]{ieeeconf}  
\IEEEoverridecommandlockouts
\usepackage{cite}
\usepackage{subfigure}
\usepackage{amsmath,amssymb,amsfonts}
\usepackage{algorithmic}
\usepackage{graphicx}
\usepackage{textcomp}
\usepackage{xcolor}
\def\BibTeX{{\rm B\kern-.05em{\sc i\kern-.025em b}\kern-.08em
    T\kern-.1667em\lower.7ex\hbox{E}\kern-.125emX}}

\title{\LARGE \bf
A Mixed Reality System for Interaction\\with Heterogeneous Robotic Systems}

\author{Valeria Villani, Beatrice Capelli and Lorenzo Sabattini
\thanks{*This work was supported by the Socially-acceptable Extended Reality Models and Systems (SERMAS) Project of the European Union’s Horizon Europe Research and Innovation Program (GA n. 101070351) and the COLLABORATION Project through the Italian Ministry of Foreign Affairs and International Cooperation.}
\thanks{The authors are with the Department of Sciences and Methods for Engineering (DISMI), University of Modena and Reggio Emilia, Reggio Emilia, Italy {\tt\small \{name.surname\}@unimore.it}}%
}

\begin{document}



\maketitle
\thispagestyle{empty}
\pagestyle{empty}

\begin{abstract}
The growing spread of robots for service and industrial purposes calls for versatile, intuitive and portable interaction approaches. In particular, in industrial environments, operators should be able to interact with robots in a fast, effective, and possibly effortless manner. To this end, reality enhancement techniques have been used to achieve efficient management and simplify interactions, in particular in manufacturing and logistics processes. Building upon this, in this paper we propose a system based on mixed reality that allows a ubiquitous interface for heterogeneous robotic systems in dynamic scenarios, where users are involved in different tasks and need to interact with different robots. By means of mixed reality, users can interact with a robot through manipulation of its virtual replica, which is always colocated with the user and is extracted when interaction is needed. The system has been tested in a simulated intralogistics setting, where different robots are present and require sporadic intervention by human operators, who are involved in other tasks. In our setting we consider the presence of drones and AGVs with different levels of autonomy, calling for different user interventions. The proposed approach has been validated in virtual reality, considering quantitative and qualitative assessment of performance and user's feedback.
\end{abstract}




\section{Introduction}\label{sec:introduction}

Driven by the advent of advanced, digital, and smart logistics and manufacturing paradigms, the use of robots and advanced automation systems has strikingly increased in recent years across all industry sectors and organization sizes. While former robots, being bulky, expensive and allowing return on investment for large production bathes, were prerogative of bigger industries, the introduction of collaborative solutions has allowed a significant spread, due to a large number of advantages mainly based on the fact that human operators and robots can work close or even together \cite{Villani_2018}. As a result, robots are no longer stand-alone systems in the factory floor \cite{Hietanen_2020, Robotics_2014_Report}. This has two implications. On the one side, factories are populated by different robots and machines, which are used to automate different operations, such as goods lifting, picking and transportation. On the other side, users frequently interact with robots, physically and cognitively, by sharing spaces and tasks.
Operators are immersed in intelligent environments and share and receive real-time information from many smart objects, such as machines, robots or even products. 

To deal with such complexity, new approaches to manage and control the manufacturing and logistics processes are required. 
To this end, natural user interfaces and interaction approaches have been explored for robot programming and information exchange between the human operator and the robot. 
Intuitive approaches refer to the use of human-friendly interaction modes, such as speech, gesture, eye tracking, facial expression, haptics, in addition to traditional keyboard, mouse, monitor, touchpad and touchscreen \cite{Neto_2010, Tsarouchi_2016, Lambrecht_2013}. Indeed, they allow users to control and program a robot by means of high-level behaviours that abstract from the robot language. As an example, speech based interaction allows to replicate human-human communication based on voice 
\cite{vanDelden_2012, Rogowski_2013, Rogowski_2012}. Generally, these approaches implement few and simple voice commands, which, however, might be still suited for simple collaborative tasks in non-complex workcells. Another approach for intuitive communication between the human and the robot relies on human action recognition and gestures 
\cite{Darvish_2018, Neto_2019, Chaudhary_2013}.

Reality enhancement and visualization techniques have also been used in this context to improve efficient management and simplify interactions in manufacturing and logistics processes \cite{Szafir_2019}. Specifically, virtual reality (VR) provides a totally immersive environment where the user's senses are under control of the system. In human-robot interaction (HRI) it is mainly used for two purposes. First, it allows to create an intuitive simulation environment where the user can interactively control a virtual model of the
robot. As a result, users can receive training and support for new collaborative tasks or workstations \cite{Loch_2017, Gavish_2015} and interaction approaches \cite{Villani_2018_ROMAN_VR}. Second, VR-based HRI allows the operator to project actions that are carried out in the virtual world onto the real world by means of robots, thus teleoperating robots in the virtual world and having them moving in the real world.
Conversely, augmented reality (AR) consists in augmenting the real world scene with virtual elements, which provide the user with additional information and enhance situational awareness, leveraging the physical interaction space as a shared canvas for visual cues.
As a result, the user spontaneously perceives the pieces of information that are useful during robot programming, task planning and  teleoperation.
Typically, head mounted displays (HMDs) are used and visual elements and information is projected directly within the user's gaze for spatial display of information. Common applications of AR for HRI are related to maintenance and assembly tasks \cite{Webel_2013, Michalos_2016}. In \cite{Andersen_2016} AR is implemented in terms of an object-aware projection technique that allows robots to visualize task information and intentions on physical objects in the environment. As a result, a human co-worker can be informed in a timely manner about the safety of the workspace, the site of next robot manipulation tasks, and next subtasks to perform. Similarly, the authors in \cite{Hietanen_2020} define a shared workspace model to monitor safety margins with a depth sensor and to communicate the margins to the operator with an interactive AR-based user interface.
AR in HRI is also used for navigation tasks with mobile robots. In \cite{Kastner_2019} AR allows to visualize navigation data, such as laser scan, environment map and path planing data. In \cite{Rosen_2019} a HMD was used for robot motion intent communication and its effectiveness was evaluated against a 2D display in a simple toy task.

It is worthwhile pointing out that AR does not provide interaction with the surrounding environment and elements, since it consists in the overlay of virtual information on the real world. Virtual elements, such as images, text and animation, augment or enhance the real world the individual is experiencing, in the sense that they passively provide additional information. Nevertheless, they are not interactive elements that can be actively used to interact with the real world. This limitation is overcome resorting to mixed reality (MR). This concept was firstly introduced by Milgram and Kishino in \cite{Milgram_1994} and refers to the possibility for virtual and real elements to interact with one another and the user to interact with virtual elements that are overlaid on the real environment. Hence, the difference of MR over AR is that the user can interact with virtual elements. 

The availability of such a mixed environment, created by the visual combination of digital content with real working spaces, has been found to be promising in HRI \cite{Krupke_2018}. First investigation about the use of MR was proposed by the authors in \cite{Hoenig_2015}, who highlighted benefits of MR in HRI as regards to spatial flexibility, elimination of safety risks, simplification of debugging and system scalability. Benefits related to simplification and enhancement of the interaction experience were, however, not considered. Nevertheless, in the following years, MR has been used for HRI. 
In particular, a mobile MR interface approach to enhance HRI in shared spaces was proposed in \cite{Frank_2017}. As the user points a mobile device at the robot's workspace, an MR environment is rendered providing a common frame of reference for the user and the robot. This allows to effectively communicate spatial information for performing object manipulation tasks, thus improving the user's situational awareness while interacting with augmented graphics, to intuitively command the robot. Pick-and-place tasks with a robot arm were considered in \cite{Krupke_2018}, where an MR system was proposed to provide experts as well as non-experts the capabilities to intuitively and naturally control a robot for selection and manipulation tasks.
MR for interactive programming of a robotic manipulator was considered in \cite{Ostanin_2020}. The system gives  the possibility to recognize the real robot location by  point cloud analysis, to use virtual markers and menus for task creation, to generate a trajectory for execution in a simulator or on the real manipulator. It also provides the possibility of scaling virtual and real worlds for more accurate planning.
Furthermore, in \cite{Rosen_2020} MR was used to provide feedback from the robot to the user in item disambiguation task, compared to deictic gestures. The authors found that MR feedback condition was more accurate and faster than the physical condition of deictic gestures.
Recently, the authors of \cite{Kennel_2023} presented an AR and MR system for interaction with multiple robots via touch gestures on a tablet or hand-tracking with a headset. Interestingly their comparison showed that users preferred touch-based interactions on a tablet for simpler tasks, while they preferred the headset for more complex interactions.

The aim of this work is to present a novel concept of MR-based HRI system that provides a ubiquitous interface for heterogeneous robotic systems.
The proposed interaction approach consists in the use of an MR tool that allows a subject to interact with a robot through manipulation of its virtual replica.
In other words, by using an MR device, the user is shown a virtual replica of the robot. Such replica is superimposed in the real scene and is meant as interaction device that enables interaction with the robot. Interaction is made possible through interacting elements, such as arrows or buttons, that come with robot virtual replica. By selecting and using such virtual elements, commands are communicated to the robot, thus leveraging virtual elements to act on real robotic agents.


\section{Proposed interaction approach}\label{sec:approach}

The proposed interaction approach is built referring to a common everyday interaction scheme, consisting in the use of a portable device that the user carries, for example, in the pocket to interact with an agent (e.g., a remote controller). When interaction is needed, the user extracts the controller from the pocket and starts communication with the agent. Building upon this idea, in the proposed approach, using an MR device, we use a virtual replica of the robot as controller for the robot and link such replica to the user's hand, so that the (virtual) interaction device is always colocated with the user. When interaction is not needed, the device is hidden to the user, as if it was in the pocket. A gateway sign, such as gesture or vocal command, is used to activate the controller, similarly to extracting it from the pocket. Once the controller is activated, the robot virtual replica becomes visible close to user's hand and interaction can be started.

\begin{figure}
	\centering
	\subfigure[Palm facing up as gateway sign to show the virtual interaction devices]{
		\includegraphics[width=.4\columnwidth]{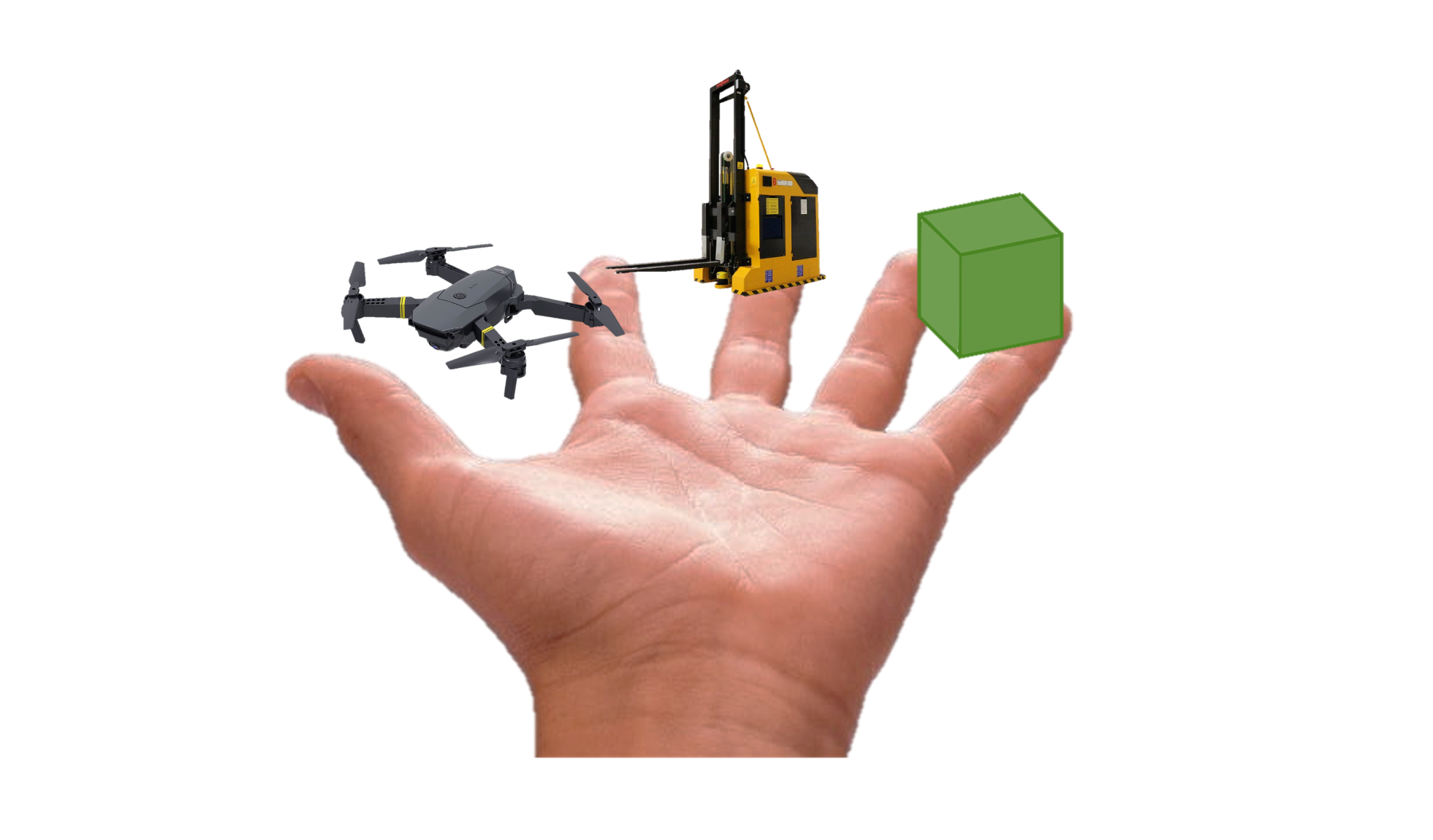}
		\label{fig:dynamic_UI1}}
	\quad
	\subfigure[Selection of an interaction device]{
		\includegraphics[width=.4\columnwidth]{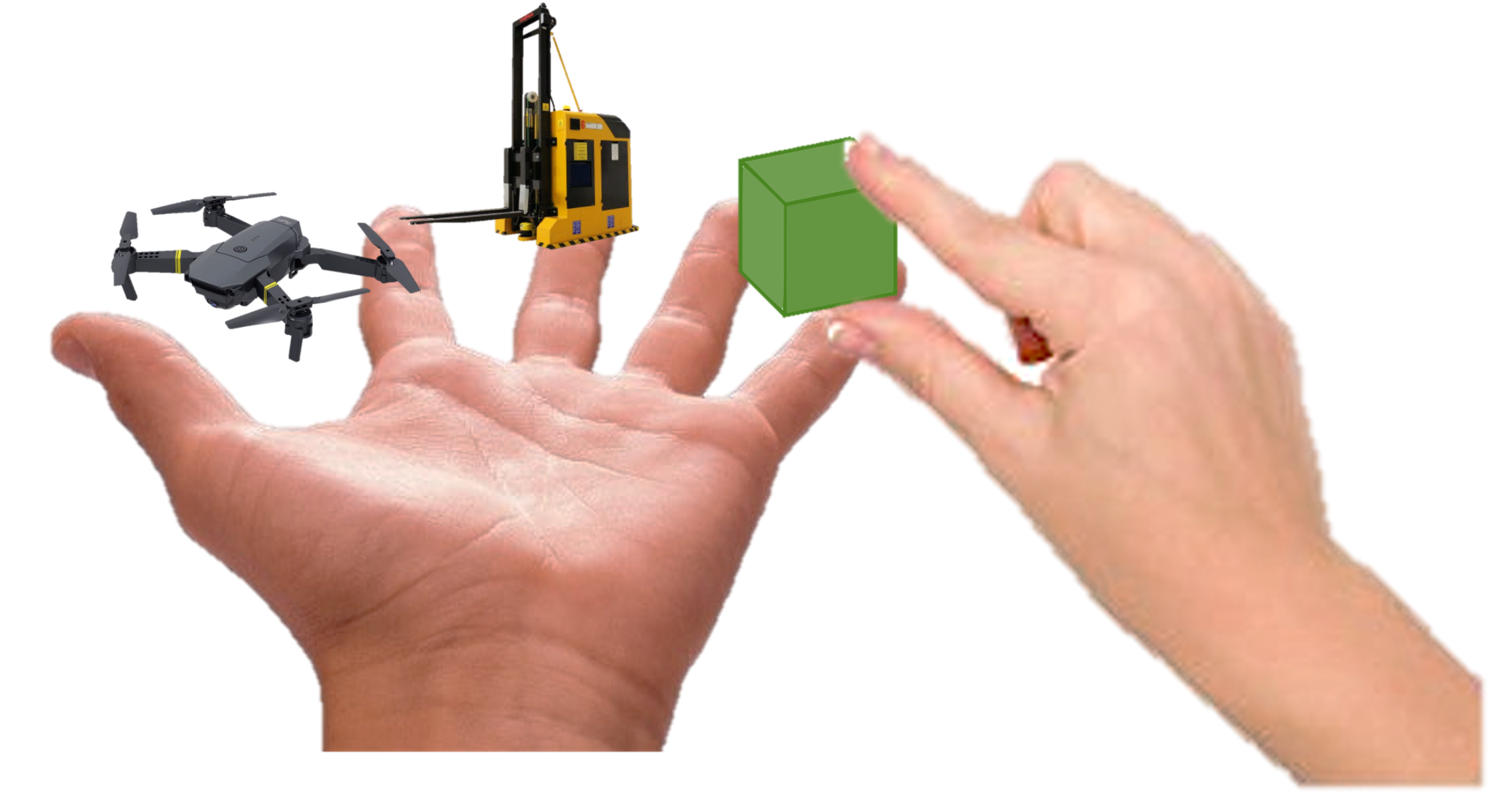}
		\label{fig:dynamic_UI2}}\\
	\subfigure[Releasing the  interaction device to open the interface]{
		\includegraphics[width=.4\columnwidth]{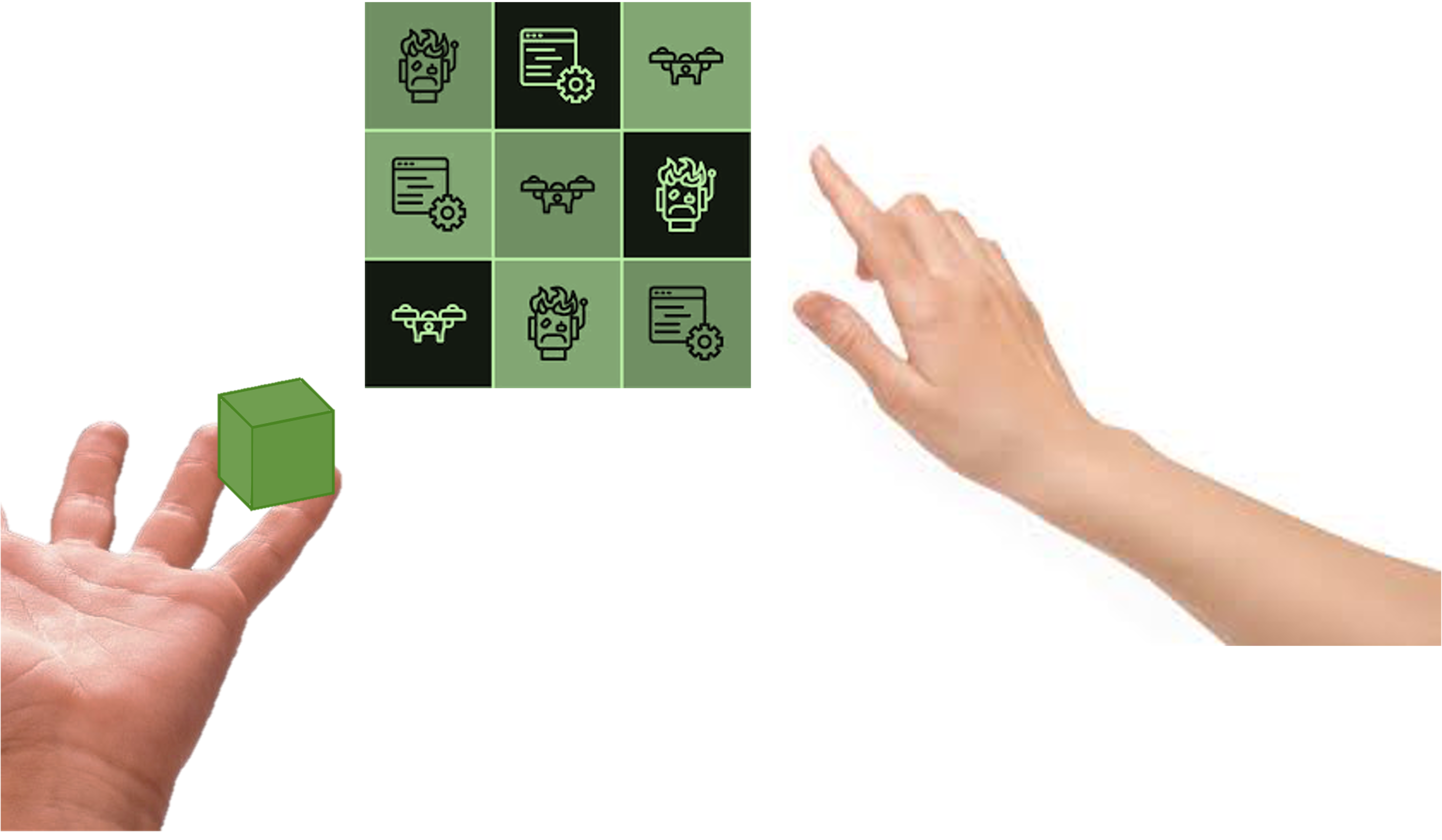}
		\label{fig:dynamic_UI3}}
	\quad
	\subfigure[Closing the interface]{
		\includegraphics[width=.4\columnwidth]{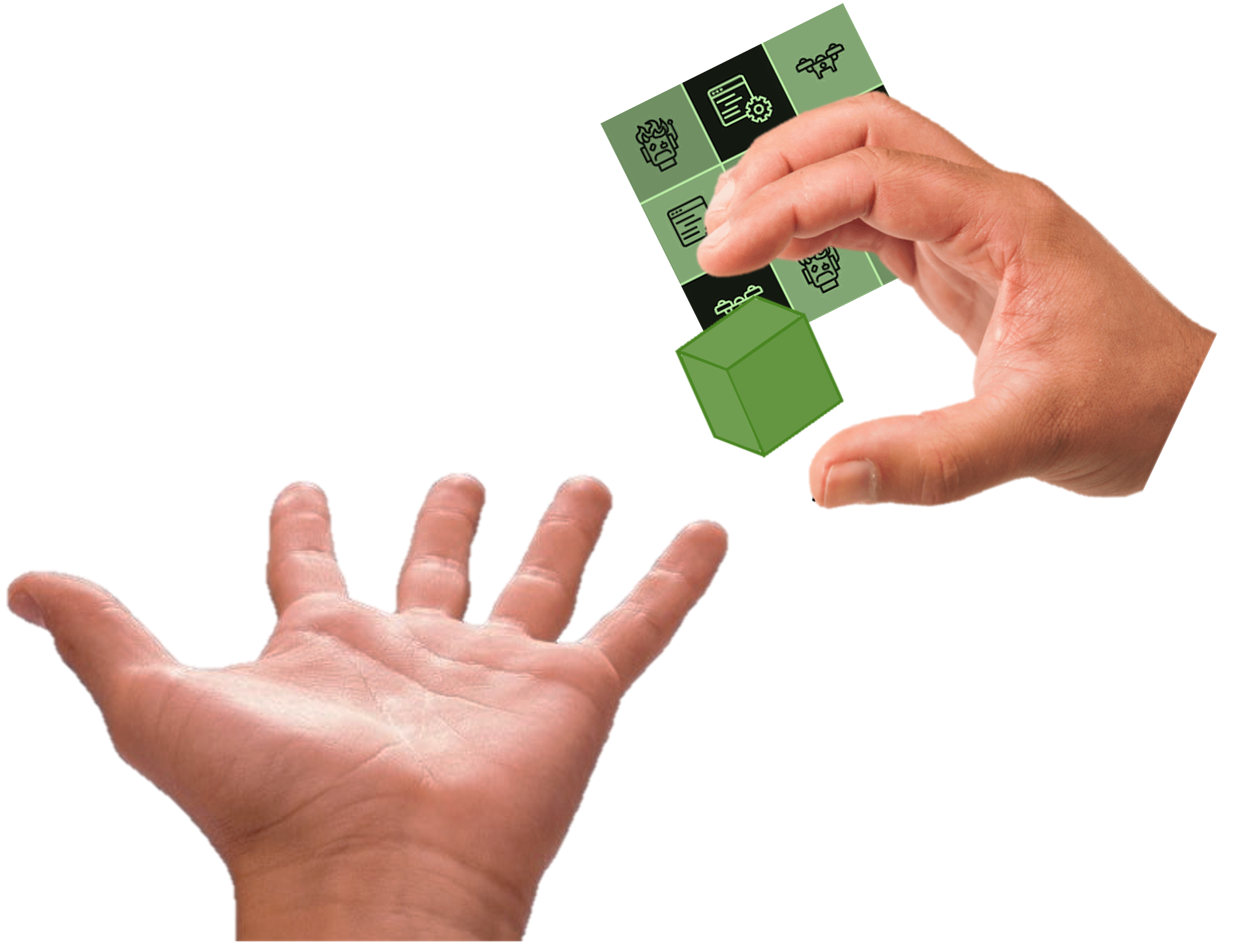}
		\label{fig:dynamic_UI4}}
	\caption{Proposed MR-based interaction approach}
	\label{fig:dynamic_UI}
\end{figure}

A reference example of the proposed interaction approach is shown in Fig.~\ref{fig:dynamic_UI}. The palm facing up, shown in Fig.~\ref{fig:dynamic_UI1}, is considered as the gateway sign that prompts the interaction device. In the figure, we consider three interaction devices, each possibly associated to a different robot. Just to provide an example, in the figure we consider the case of a drone, an automated guided vehicle (AGV) and another robot, whose replica is represented by a cube, for the sake of generality.
To start the interaction with a given robot, the user is expected to expand the corresponding virtual replica, that is to grab and release it (Fig.~\ref{fig:dynamic_UI2}), as if a physical controller was kept in its case or was turned off. Once the (physical) case is opened or the (virtual) controller is released, the interaction panel is visible (Fig.~\ref{fig:dynamic_UI3}) and the user can start commanding the robot. The interaction panel contains virtual interaction devices, such as arrows, buttons or displays, that allow exchange of commands with the robot.
At the end of the interaction task, the controller can be put away by taking it to its original position (Fig.~\ref{fig:dynamic_UI4}), as one would do in the pocket with a physical device.

The example shown in Fig.~\ref{fig:dynamic_UI} is general and will be instantiated in Subsec.~\ref{subsubsec:interfacce_drone_AGV} to the case of drones and AGVs.

A first straightforward advantage of the proposed approach comes from the fact that no physical device is needed to interact with robots, apart from MR goggles. Moreover, since the interaction elements are digital and virtual, great flexibility and versatility can be achieved within a single interaction approach. In particular, this implies that the need for multiple devices controlling multiple robots is eliminated, since different robots can be selectively associated to different virtual interaction elements. 
To this end, virtual interaction elements, such as arrows, can be added as affordances to robots, since they represent a prompt on what can be done with the robot. Indeed, they allow a clear mapping between user's possible or exerted actions and directional commands to the robot. For example, constrained motion directions can be made visible by showing interaction elements related to admitted directions.
Additionally, in the case of robots implementing predefined behaviors, the use of specific digital affordances and signifiers, which are used to indicate an affordance, makes the interaction more intuitive and easier. Digital buttons can be used to start such behaviors or, in the case of a robot mounting specific tools, digital avatars of the tools allow easy recall for the user and the tools can be easily activated or selected by interacting with the corresponding avatars.

\subsection{Working Scenario}\label{subsec:scenario}
To instantiate the proposed interaction approach, we consider, without any lack of generalization, an intralogistics setting, where goods have to be moved in company premises and warehouse. 
In particular, we consider an heterogeneous human-multi-robot system, which includes drones and AGVs. Different levels of autonomy might be available for these robots, ranging from fully automated driverless transport to manually guided solutions, depending on available infrastructure and onboard sensors \cite{Morth_2020}. In the following, we consider the case of partially automated robots, which, hence, call for human operators interventions and require human-robot interfaces.

As regards drones, we envision a scenario where they are used to lift and move goods over long distances in the warehouse. The rationale behind this is a large warehouse where operators work in restricted areas but goods need to be moved from an area to another (for example, from long-term stocking to kitting). A drone is then used to move goods across different areas, in order to limit operator's displacements over long distances.
Fixed positions for taking off and landing, where to grasp and release goods to be carried, can be considered, but different orientations of goods at take off position can be allowed to avoid the need for an accurate deposit by operators.
Additionally, fleets of AGVs can be used to operate autonomously in the environment shared with human operators, since AGVs can be charged of repetitive tasks across the warehouse. Interaction with human operators might be required when new tasks or displacements have to be programmed. In this case, it is likely that possible routes for AGVs are preprogrammed and operators can choose among them, or program new ones.

\subsubsection{Interfaces for drone and AGVs interaction}\label{subsubsec:interfacce_drone_AGV}
As in Fig.~\ref{fig:dynamic_UI}, an avatar for each robot is created as interaction device. Each avatar resembles the associated robot, in order to facilitate mapping between digital and real worlds. Moreover, while robots move, the corresponding avatars move with them.

\begin{figure}
	\centering
	\includegraphics[width=.6\columnwidth]{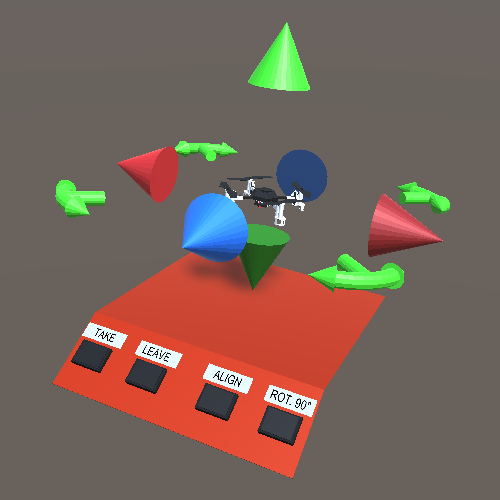}
	\caption{The mixed reality interface to control a drone}
	\label{fig:interfaccia_drone}
\end{figure}

To enable the interaction with drones, the corresponding interaction panel, which is activated by selecting the corresponding avatar, is composed by three parts, as shown in Fig.~\ref{fig:interfaccia_drone}: robot avatar, a set of arrows and a button panel. In the middle, the robot avatar, namely a replica of the robot, is shown, to facilitate association between commands and robot to command. Arrows allow to command robot movements, while the button panel enables commands related to object picking and placing and shortcuts for predefined movements. These interfaces are dynamically adapted according to the current state of the system. 

In particular, arrows determine drone translations and rotations. Once grabbed by the user, they allow translations along the three Cartesian axes centered in the drone and rotations around its longitudinal axis. Different shapes and colors are used to make selection of the desired arrow intuitive. All the arrows 
become visible when user's hands get in proximity to the robot, otherwise they are hidden, in order to prevent occlusions in user's field of view.
As regards buttons in the panel, they are used to grasp and release the items to be picked. Additional buttons can be added to command common preprogrammed actions for the robot, such as predefined rotations to facilitate alignment with the items to be picked and, hence, picking. Nevertheless, these actions are needed not at the same time. Thus, to avoid confusion in the user, the corresponding buttons are selectively activated depending on robot state.

\begin{figure}
	\centering
	\includegraphics[width=\columnwidth]{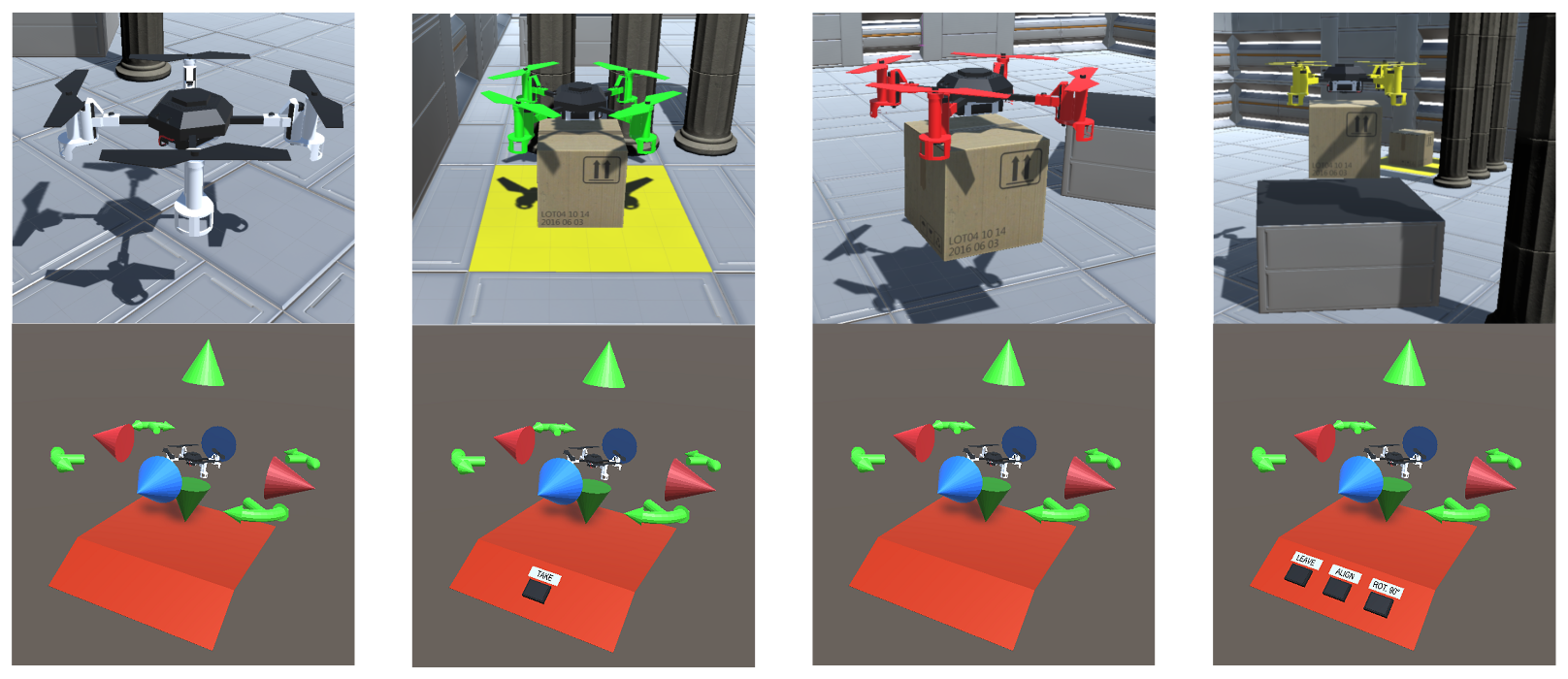}
	\caption{Different states of the interface. From left to right: (1) Drone in freedrive state; (2) Drone ready to pick the box; (3) Drone picking the box; (4) Drone ready to release the box}
	\label{fig:interfaccia_drone_stati}
\end{figure}

Four drone states have been identified to this end:
\begin{enumerate}
	\item \textit{freedrive}: this applies to all the time instants when the drone is idle and is not involved in any specific action; in this state the only possible action is to move it: hence, no buttons are shown and the user can move the robot with the arrows;
	\item \textit{ready to pick}: when in this state, the robot is ready to grasp objects; thus, in the button panel, the ``Grasp'' option is selected;
	\item \textit{picking}: this state refers to the robot carrying an object and flying to the landing position. In this state, no other actions can be commanded by the user; hence, no buttons are shown in the panel. If robot flight is not autonomous, arrows stay visible to let the user move the robot;
	\item \textit{ready to release}: this state refers to the drone in the landing position, ready to release the carried object. Before release, it might be useful to change robot orientation, in order to align the object to the landing position (e.g., align boxes to stack). To this end, in this state the button panel shows three buttons:  ``Release'', ``Rotate $90$\textdegree'' and, if a vision system is available onboard the robot, ``Align''.
\end{enumerate}
These states are also notified with a different color for the robot avatar: dark grey for the default \textit{freedrive} state, green for \textit{ready to pick}, red for \textit{picking} and yellow for \textit{ready to release}. The interfaces corresponding to these states are shown in Fig.~\ref{fig:interfaccia_drone_stati}.

Finally, if the robot flight needs to be teleoperated and a collision detection system is not available onboard, the user, upon request, can be shown the view of the camera onboard the drone. 
This allows to teleoperate the robot while being aware of where it is moving. This feature can be activated and deactivated with a hand gesture, such as thumb up in our experiments.

As regards the interaction with AGVs, the corresponding interaction panel is simplified since possible motions are reduced with respect to drones. In particular, considering a nonholonomic AGV, arrows for moving backwards and forwards are the only available, together with those for rotations around the longitudinal axis. If robot routes are preprogrammed, the button panel can be customized to command them. {Other possible actions to be commanded are lifting forks for vehicle loading and unloading or commanding the robot to reach charging position.}


\section{Experimental Methods}\label{sec:methods}

We designed an experiment to assess the proposed interaction approach in the considered working scenario related to intralogistics applications.
 
The proposed approach has been compared to the use of standard interaction devices commonly used in this scenario, such as joypads. Robot-specific joypads have been considered, that is one for drones and a different one for AGVs.

\subsection{Experimental Environment}\label{subsec:environment}

\begin{figure}
	\centering
	\includegraphics[width=.8\columnwidth]{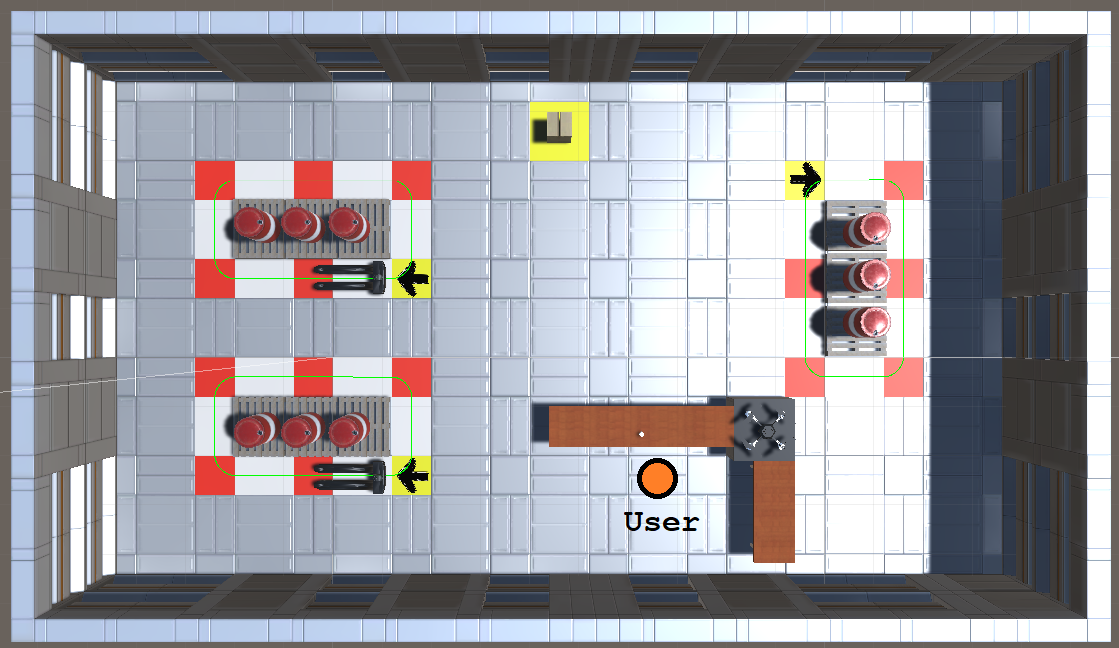}
	\caption{Experimental scenario}
	\label{fig:scena}
\end{figure}

The working scenario presented in Sec.~\ref{subsec:scenario} was implemented considering the environment depicted in Fig.~\ref{fig:scena}, which includes all the items mentioned above. In particular, two AGVs have been placed in the left part of the scene and the red and white paths show their current trajectories. On the right, another possible trajectory for AGVs is depicted. In the middle, the working area of a human operator is shown, with two working tables. Close to the working tables, a drone has been placed in its take off location, while the yellow square represents the landing spot.

We created this working scenario in a VR environment, in order to have an immersive and reliable replica of a realistic intralogistics setting. Hence, Fig.~\ref{fig:scena} represents the top view of the scene available to test subjects. Successful use of virtual reality for the evaluation of HRI systems was explored in \cite{Villani_ROMAN_2018_VR}.

\subsection{Experimental Task}\label{subsec:task}
With reference to the scenario described in Sec.~\ref{subsec:scenario}, test participants were asked to perform the same experimental task twice, considering two different interaction approaches. In the following, we name experimental session each part of the experiment carried out with each interaction approach: in other words, the entire experiment consisted of two experimental sessions, one with the proposed MR-based interaction and the other with the joypad.
Users could find the joypad in a place located close to the working tables where they were standing.
	
Within each session, we introduced a primary and a secondary task: the former accounts for manual activities in charge solely to the human operator, while the latter refers to occasional interaction with the robots in the scene. For the primary task, as a toy example of activities in charge to operators, we consider the need to move boxes from one side to the other of the working table: this can represent an assembly or kitting task, in a real scenario. As regards the secondary task, it represents an interruption of the primary one and is randomly presented to the test subject. Given the presence of three different robots in the scene in Fig.~\ref{fig:scena}, two different secondary tasks are possible:

\begin{enumerate}
	\item guide one of the AGVs to the beginning of the path on the right (black arrow on yellow background on the right) and command it to follow the corresponding route;
	\item command the drone to lift a box and move it from take off to land position. 
\end{enumerate}

{The occurrence of a new secondary task is notified differently depending on the interaction modality: an alert is shown on a screen placed on the working table in the case of the joypad and in the MR device in the case of MR-based interaction.}

The overall duration of the experiment was, on average, of 3.5 minutes approximately. At the beginning of each experimental session, test subjects were presented with the primary task. After 30 seconds, the need to carry out one of the secondary tasks with the robots was notified. When the subject completed the secondary task, they were asked to continue with the primary task and, after further 30 seconds, they were notified for the other secondary task. After completion of the second secondary task, the experimental session was concluded. 


For each experiment, the subject was asked to read a description of the experiment and sign a consent form. After signing the consent form, the experimental protocol was explained to the subject, demographic information was collected and a tutorial was shown to let the subject get acquainted with the VR setup. At the end of each session, a usability questionnaire was administered to test participant.

\subsection{Test Subjects}\label{subsec:subjects}

A total of $N = 24$ users ($6$ females, $18$ males, mean age \mbox{$24.6$ y.o.}, \mbox{$21$ y.o.} the youngest, \mbox{$43$ y.o.} the oldest) were enrolled in the experiments. Participants are researchers working at our engineering department. All of them were completely new to the experimental task and goals.

According to a within-subject design~\cite{montgomery2017design}, test subjects were exposed to the two interaction modalities (MR-based and joypad), and to both tasks within each modality. In addition, the order in which the subjects tested the interaction modalities and the tasks was imposed by a Latin rectangle design~\cite{montgomery2017design}, in such a way to reduce the residual error introduced by the nuisance effect of learning effect.

\subsection{Metrics}\label{subsec:metrics}

To compare the two interaction modalities included in the experiment, we considered quantitative and qualitative assessment of performance and user's feedback. As regards quantitative metrics, for each experimental session we measured the duration of the following time intervals:

\begin{itemize}
	\item \textit{total time}: it represents the duration of the session, measured from the instant when the first secondary task is notified to the completion of the second secondary task. It provides an overall comparison of the two HRI modalities.

	\item \textit{robot time}: it is the time required to command the robot,  measured from the instant when the MR-based interaction is extracted or the joypad is activated to the completion of the secondary task. It is used to assess the efficiency of MR-based interaction with respect to the joypad to interact with robots.
	
	\item \textit{reaction time}: it accounts for the time required to activate the interaction system, that is to extract the MR-based interaction panel or activate the joypad. It is measured starting from the time instant when the incoming secondary task is notified; it focuses on the assessment of the modality considered for notifying the need of user intervention.
\end{itemize}

As regards subjective measurements, a questionnaire was administered to collect feedback about interaction usability and user's satisfaction. 
The questionnaire was organized in three parts. The first was administered at the end of the first experimental session and consists in the System Usability Scale (SUS) \cite{Brooke_1996} related to the interaction modality tested in the first session: this is an established tool for measuring the usability of a wide variety of products and services and represents a short, simple tool, practical to be used in industrial settings \cite{Villani_2021}. Moreover, it was shown that the SUS is a highly robust and versatile tool for usability professionals \cite{Bangor_2008}. Then, at the end of the second experimental session, a SUS questionnaire referring to the other interaction modality was administered. Finally, we added three direct comparative questions asking what users preferred between MR-based interaction and joypad in general and specifically for interacting with drones and AGVs. The questions were ``\textit{Which interaction approach have you preferred?}'', ``\textit{Which approach have you preferred for interacting with the drone?}'' and ``\textit{Which approach have you preferred for interacting with the AGVs?}''.
Blank space for open comments regarding this last set of questions was provided.

As a result, 23 statements were included in the questionnaire. User's agreement with SUS statements was collected with a $5$-point Likert scale, where $1$ = \textit{strongly disagree} and $5$ = \textit{strongly agree}.
The entire questionnaire was integrated in the VR system, to provide continuity in the immersive experience.

\subsection{Implementation}\label{subsec:implementation}

The scenario of VR was developed in Unity (Unity Technologies, California, USA), a free game development platform with a built-in physics and rendering engine. The Oculus Rift headset (Oculus VR, USA) was used for immersive VR. For MR-based interaction, hand tracking was achieved with the Leap Motion Controller (Ultraleap, California, USA), and the framework was developed using the Unity Modules Package supplied by Ultraleap\footnote{Avaible at: https://github.com/leapmotion/UnityModules.}. A common wireless joypad for gaming (Logitech Gamepad F710) was used for comparison with MR-based interaction.

\section{Results}\label{sec:results}

\subsection{Time performance}\label{subsec:results-tempi}

\begin{figure}
	\centering
	\includegraphics[width=\columnwidth]{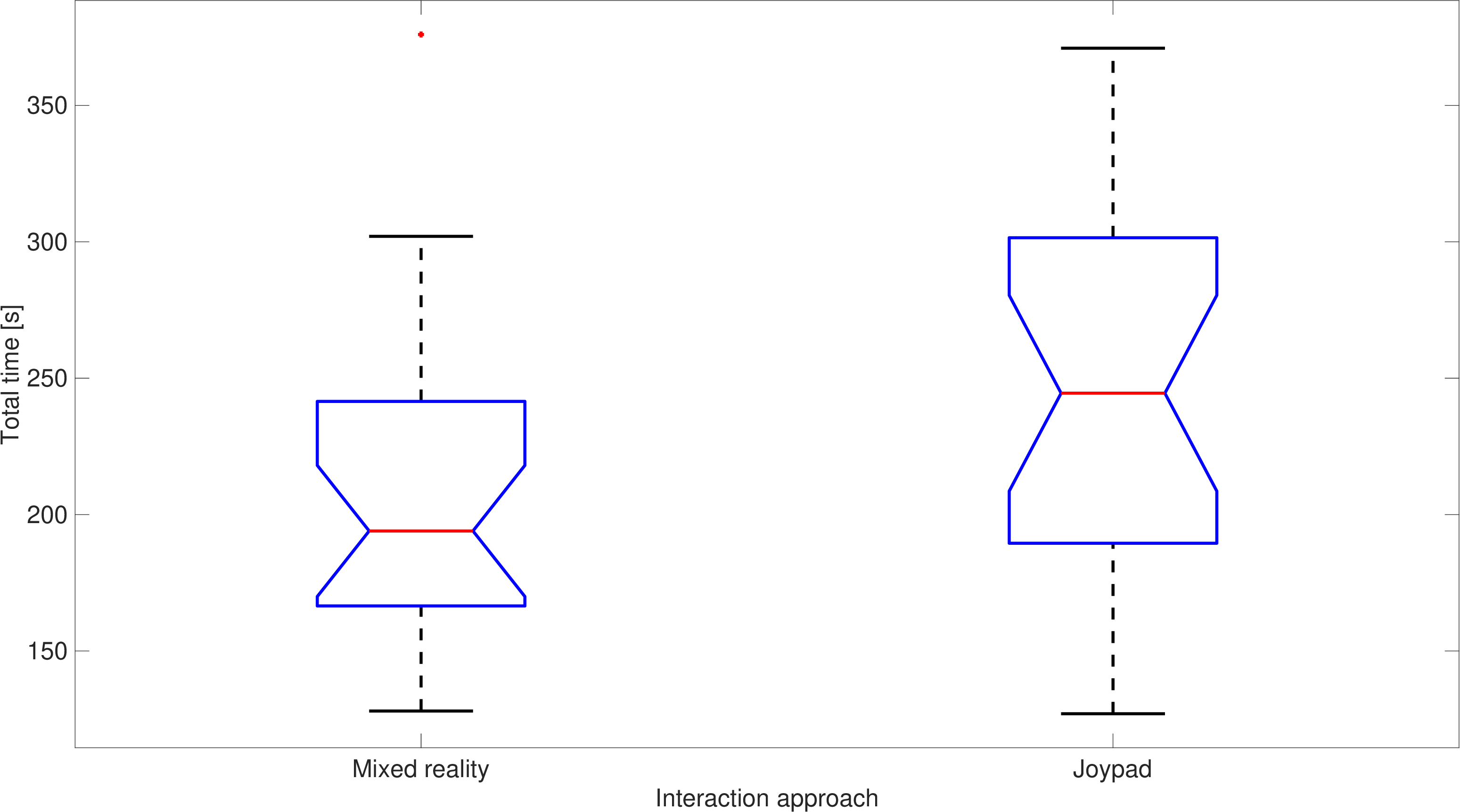}
	\caption{Experimental results: total time}
	\label{fig:total_time}
\end{figure}

Figure~\ref{fig:total_time} reports the \textit{total time} of the experimental sessions with MR-based interaction and the joypad. Results show that the experiment could be completed more quickly when using the MR system. In particular, \textit{total time} was measured as $208.7\pm58.5~\text{s}$ with the MR-based approach and $245.2\pm73.7~\text{s}$ with the joypads. Applying the t-test, the difference proved statistically significant with $p=0.0274$. The paired sample t-test is used for understanding if the mean difference between two sets of observations is zero, i.e., in this case whether the modality of interaction affects the performance of the users. We have paired data as each subject was tested with both modalities.

\begin{figure}
	\centering
    \includegraphics[width=\columnwidth]{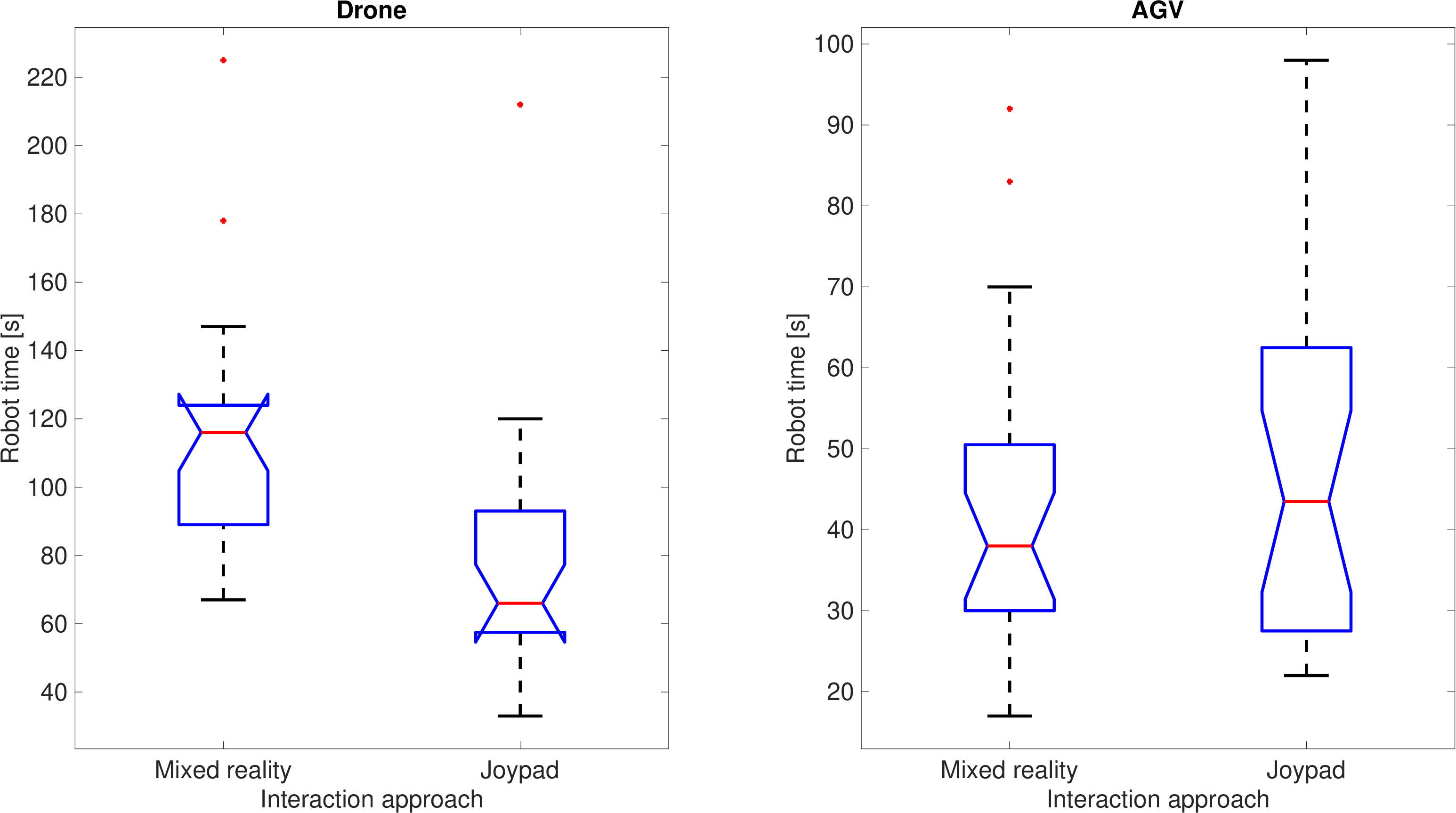}
    \caption{Experimental results: robot time}	\label{fig:robot_time}
\end{figure}

\begin{figure}
	\centering
	\includegraphics[width=\columnwidth]{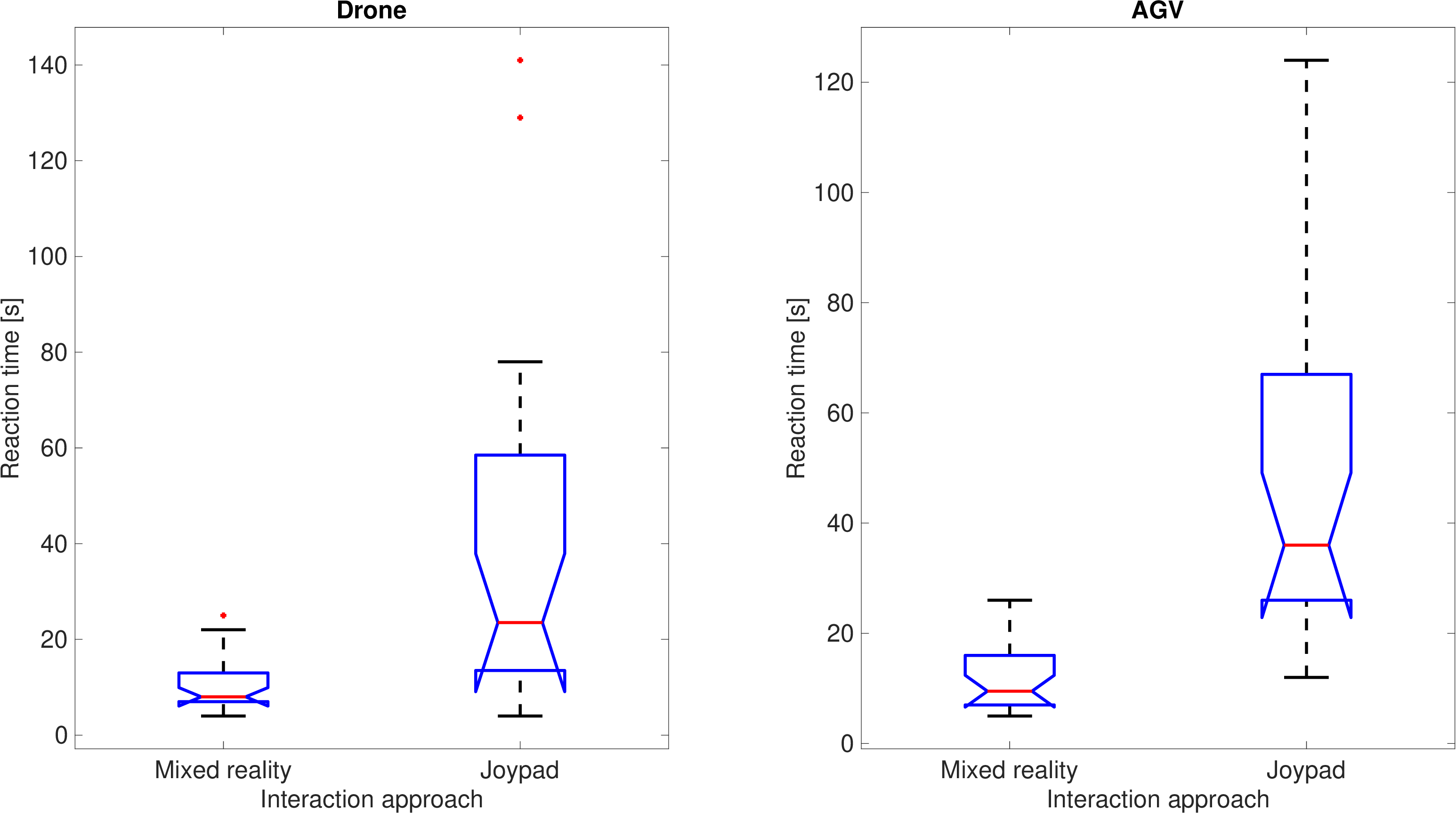}
	\caption{Experimental results: reaction time}
	\label{fig:reaction_time}
\end{figure}

\begin{figure*}
	\centering
	\includegraphics[width=.8\textwidth]{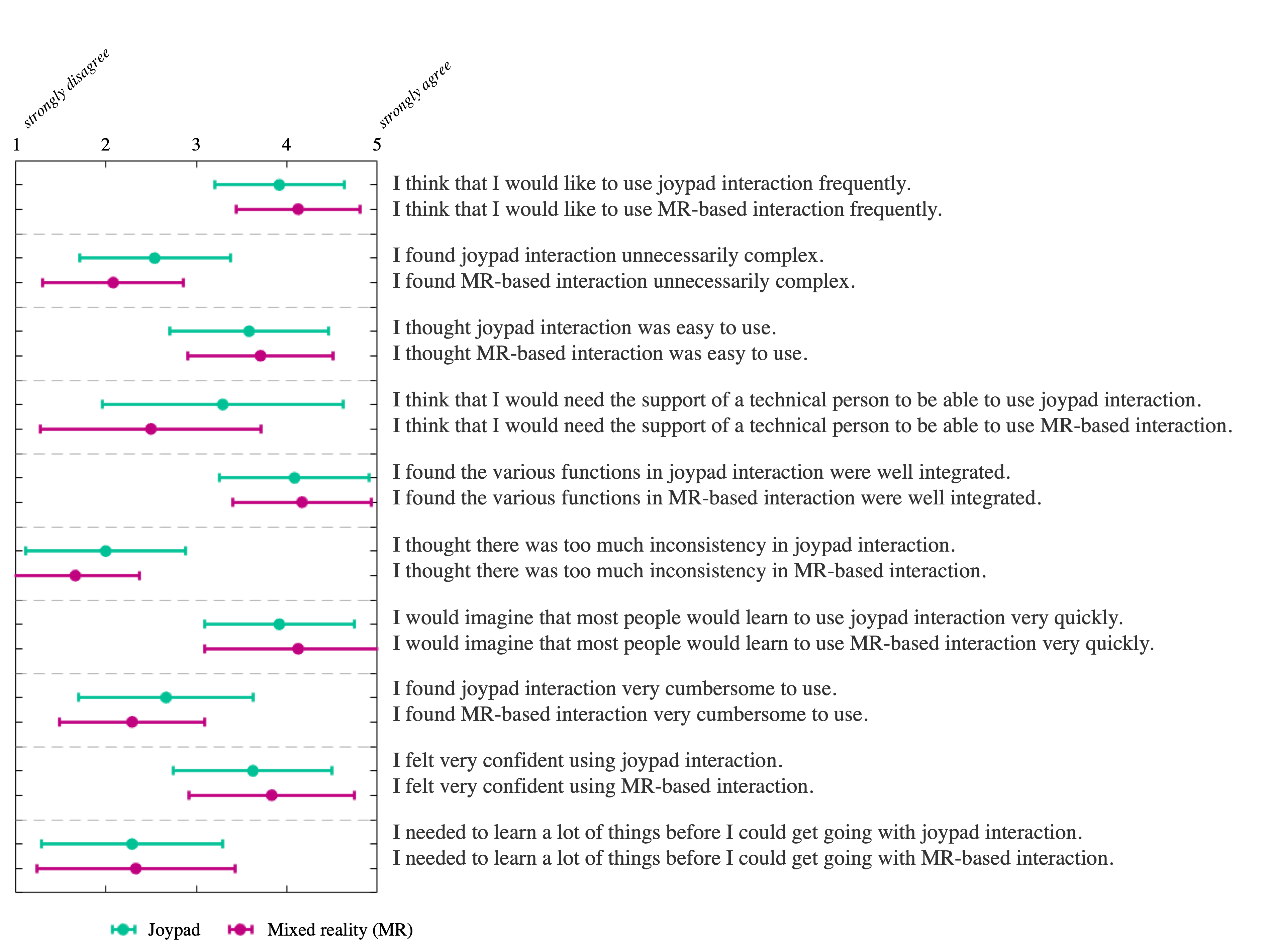}
	\caption{Replies to SUS questionnaires investigating interaction with joypad and the proposed MR-based approach}
	\label{fig:SUS}
\end{figure*}

Considering \textit{robot time}, that is the time required to command the robot, from the activation of the MR-based interaction panel or joypad to the end of the task, the use of the drone and AGVs reported different results, which are summarized in Fig.~\ref{fig:robot_time}. Specifically, as regards the drone, differences have been found between the two interaction modalities: using the MR-based approach resulted slower than the joypad, with the difference being statistically significant (MR-based: $113.3\pm36.5~\text{s}$, joypads: $78.0\pm36.9~\text{s}$, $p=0.0035$). As regards the AGVs, difference between the two approaches is slightly in favor of the MR-based one, although not statistically significant (MR-based: $42.5\pm19.6~\text{s}$, joypads: $49.7\pm24.0~\text{s}$, $p=0.1305$).

Finally, Fig.~\ref{fig:reaction_time} reports the measured \textit{reaction times}. The results were compared with a t-test considering the interaction modality as only independent variable. \textit{Reaction time} when an incoming task is notified was statistically significantly lower when MR-based interaction was used ($p-value < 0.001$).


All together, these results prove that differences measured for \textit{total times} are due to the different way used to notify the user when an action was requested to deal with an incoming secondary task, rather than to the way commands are communicated to robots. On the contrary, the use of MR resulted disadvantageous for \textit{robot time}, at least in the case of drone. Likely, this was due to the lack of familiarity that test subjects had with MR. It is then possible to conclude that the greatest advantages brought by MR-based interaction are due to increased user's responsiveness to an incoming task and the fact that the interaction device is colocated with the user.


\subsection{Subjective assessment with questionnaires}\label{subsec:results-questionari}

Figure~\ref{fig:SUS} reports the replies to the SUS questionnaires administered after each experimental session, with either MR-based interaction or joypads. The figure shows that perceived usability was almost similar for the two interaction methods. The only relevant difference has been reported with respect to the need of support by a technical person to use the two systems: average replies scored $2.50\pm1.22$ for MR-based interaction and $3.29\pm1.33$ for joypads. We believe that this is due to the fact that, thanks to virtual avatars and precise mapping between digital elements and real world commands, users have all needed information and explications when using the MR-based approach. On the contrary, when using joypads, remarkable effort is needed to remember actions associated to buttons and how to command a desired displacement to the robot.
This is, at least in part, confirmed also by replies to the last SUS statement, investigating whether it was needed to learn a lot of things before being able to use the system. Indeed, while all the test participants had previous experience with the use of joypads, they were using the proposed approach for the first time ever. Nevertheless, 15 out of 24 subjects replied $1$ (\textit{strongly disagree}) or $2$ (\textit{disagree}) to this statement referred to MR-based interaction. As well, most subjects stated that they would imagine that most people would learn to use the proposed approach very quickly: $20$ out of $24$ replied $4$ (\textit{agree}) or $5$ (\textit{strongly agree}) to this statement.

As regards the replies to the last set of questions, $62.5\%$ preferred MR-based interaction over joypads, in general (first question); nevertheless, only $45.8\%$ and $29.2\%$ preferred it when it comes to interacting with the drone and the AGVs, respectively (second and third questions). Open comments to these questions explained that the overall design of the MR-based approach was greatly appreciated since it guides users and keeps them aware about the co-presence of robots; nevertheless, previous familiarity with joypads made users more confident to use them to provide specific commands to the robots, as shown also by measured \textit{robot times} discussed in Fig.~\ref{fig:robot_time}.

\section{Conclusions}\label{sec:conclusions}

In this paper we proposed a novel approach for letting a human interact with a robot. Based on the concept of MR, the proposed interaction system allows a user to interact with a robot through intuitive manipulation of its virtual replica. 
In particular, a general purpose system was developed, that allows a user to interact with different kinds of mobile robots, adapting to their kinematic constraints. 
The proposed method was validated by means of user studies conducted in a VR environment. In such studies, we collected quantitative results related to the performance, as well as subjective results related to the usability. The analysis of the results shows that the proposed methodology represents a valid tool that can be effectively used in several kinds of applications.

\section*{Acknowledgement}
The authors would like to thank Luca Cristuib Grizzi for providing support in the development of the proposed interaction system.

\bibliographystyle{IEEEtran}
\bibliography{biblio}

\end{document}